\documentclass[preprint,5p,12pt,twocolumn]{article}

\usepackage{amssymb}
\usepackage{subfig}
\usepackage{algorithm,algorithmic}
\usepackage{graphicx}
\usepackage{caption}
\usepackage{xspace}
\usepackage{multirow}
\usepackage{rotating}
 \usepackage{url}
\usepackage{gensymb}
\usepackage [ table ]{ xcolor }
\usepackage[top=2cm, bottom=2cm, left=1.5cm, right=1.5cm]{geometry}
\usepackage{setspace}
\usepackage{morefloats}

\usepackage[affil-it]{authblk}
\usepackage{hyperref}

\usepackage[absolute]{textpos}
\begin{document}

\begin{textblock}{30}(1.5,0.5)
\noindent{\large  \textcolor{red}{\textbf{This manuscript is published in Neural Networks. Please cite it as:}}}\\
\textit{\textcolor{blue}{Kheradpisheh, S.R., Ganjtabesh, M., Thorpe, S.J., Masquelier, T.,
STDP-based spiking deep \\convolutional neural networks for object recognition. Neural Networks (2017).\\ \url{https://doi.org/10.1016/j.neunet.2017.12.005}} }
\end{textblock}

\title{STDP-based spiking deep convolutional neural networks\\ for object recognition}

\author{Saeed Reza Kheradpisheh$ ^{1,2,}$\footnote{Corresponding author.\\ Email addresses:\\ kheradpisheh@ut.ac.ir (SRK), \\ mgtabesh@ut.ac.ir (MG),\\ simon.thorpe@cnrs.fr (ST)\\ timothee.masquelier@cnrs.fr (TM).} }
\author{Mohammad Ganjtabesh$ ^{1}$}
\author{Simon J. Thorpe$ ^{2}$}
\author{Timoth\'ee Masquelier$ ^{2}$}
\affil{\footnotesize $ ^{1} $ Department of Computer Science, School of Mathematics, Statistics, and Computer Science, University of Tehran, Tehran, Iran}
\affil{\footnotesize $ ^{2} $ CerCo UMR 5549, CNRS -- Universit\'e Toulouse 3, France}

\date{}

\maketitle
\begin{abstract}
Previous studies have shown that spike-timing-dependent plasticity (STDP) can be used in spiking neural networks (SNN) to extract visual features of low or intermediate complexity in an unsupervised manner. These studies, however, used relatively shallow architectures, and only one layer was trainable. Another line of research has demonstrated -- using rate-based neural networks trained with back-propagation -- that having many layers increases the recognition robustness, an approach known as deep learning. We thus designed a deep SNN, comprising several convolutional (trainable with STDP) and pooling layers. We used a temporal coding scheme where the most strongly activated neurons fire first, and less activated neurons fire later or not at all.  The network was exposed to natural images. Thanks to STDP, neurons progressively learned features corresponding to prototypical patterns that were both salient and frequent. Only a few tens of examples per category were required and no label was needed. After learning, the complexity of the extracted features increased along the hierarchy, from edge detectors in the first layer to object prototypes in the last layer. Coding was very sparse, with only a few thousands spikes per image, and in some cases the object category could be reasonably well inferred from the activity of a single higher-order neuron. More generally, the activity of a few hundreds of such neurons contained robust category information, as demonstrated using a classifier on Caltech 101, ETH-80, and MNIST databases. We also demonstrate the superiority of STDP over other unsupervised techniques such as random crops (HMAX) or auto-encoders. Taken together, our results suggest that the combination of STDP with latency coding may be a key to understanding the way that the primate visual system learns, its remarkable processing speed and its low energy consumption. These mechanisms are also interesting for artificial vision systems, particularly for hardware solutions.\\
\textit{\textbf{Keywords:} Spiking Neural Network, STDP, Deep Learning, Object Recognition, and Temporal Coding}
\end{abstract}



\section*{Introduction}
Primate's visual system solves the object recognition task through  hierarchical processing along the ventral pathway of the visual cortex~\cite{dicarlo2012does}. Through this hierarchy, the visual preference of neurons gradually increases from oriented bars in primary visual cortex (V1) to complex objects in inferotemporal cortex (IT), where neural activity provides a robust, invariant, and linearly-separable object representation~\cite{dicarlo2012does,dicarlo2007untangling}. Despite the extensive feedback connections in the visual cortex, the first feed-forward wave of spikes in IT ($\sim 100-150$ ms  post-stimulus presentation) appears to be sufficient for crude object recognition~\cite{thorpe1996speed,hung2005fast,liu2009timing}. 

During the last decades, various computational models have been proposed to mimic this  hierarchical feed-forward processing~\cite{Fukushima1980,LeCun1998,Serre2007.PAMI,masquelier2007unsupervised,Lee2009}. Despite the limited successes of the early models~\cite{pinto2011comparing,ghodrati2014feedforward}, recent advances in deep convolutional neural networks (DCNN) led to high performing models~\cite{Krizhevsky2012,zeiler2014visualizing,simonyan2014very}. Beyond the high precision, DCNNs can tolerate object variations as humans do~\cite{kheradpisheh2015deep,kheradpisheh2016humans}, use IT-like object representations~\cite{cadieu2014deep,khaligh2014deep}, and match the spatio-temporal dynamics of the ventral visual pathway~\cite{cichy2016comparison}.

Although the architecture of DCNNs is somehow inspired by the primate's visual system~\cite{lecun2015deep} (a hierarchy of computational layers with gradually increasing receptive fields), they totally neglect the actual neural processing and learning mechanisms in the cortex. 

The computing units of DCNNs send floating-point values to each other which correspond to their activation level, while, biological neurons communicate to each other by sending electrical impulses (i.e., spikes). The amplitude and duration of all spikes are almost the same, so they are fully characterized by their emission time. Interestingly, mean spike rates are very low in the primate visual systems (perhaps only a few of hertz~\cite{shoham2006silent}). Hence, neurons appear to fire a spike only when they have to send an important message, and some information can be encoded in their spike times. Such spike-time coding leads to a fast and extremely energy-efficient neural computation in the brain (the whole human brain consumes only about 10-20 Watts of energy~\cite{Maass2002}). 

The current top-performing DCNNs are trained with the supervised back-propagation algorithm which has no biological root. Although it works well in terms of accuracy, the convergence is rather slow because of the credit assignment problem~\cite{edmund2002computational}. Furthermore, given that DCNNs typically have millions of free parameters, millions of labeled examples are needed to avoid over-fitting.  However, primates, especially humans, can learn from far fewer examples while most of the time no label is available. They may be able to do so thanks to spike-timing-dependent plasticity (STDP), an unsupervised learning mechanism which occurs in mammalian visual cortex~\cite{meliza2006receptive,huang2014associative,mcmahon2012stimulus}. According to STDP, synapses through which a presynaptic spike arrived before (respectively after) a postsynaptic one are reinforced (respectively depressed). 

To date, various spiking neural networks (SNN) have been  proposed to solve object recognition tasks. A group of these networks are actually the converted versions of traditional DCNNs~\cite{cao2015spiking,hunsberger2015spiking,diehl2016conversion}. The main idea is to replace each DCNN computing unit with a spiking neuron whose firing rate is correlated with the output of that unit.  The aim of these networks is to reduce the energy consumption in DCNNs. However, the inevitable drawbacks of such spike-rate coding are the need for many spikes per image and the long processing time. Besides, the use of back-propagation learning algorithm and having both positive (excitatory) and negative (inhibitory) output synapses in a neuron are not biologically plausible. On the other hand, there are SNNs which are originally spiking networks and learn spike patterns. First group of these networks exploit learning methods such as auto-encoder~\cite{panda2016unsupervised,burbank2015mirrored} and back-propagation\cite{bengio2015towards} which are not biologically plausible. The second group consists of SNNs with bioinspired learning rules which have shallow architectures~\cite{brader2007learning,habenschuss2012homeostatic,querlioz2013immunity,zhao2015feedforward,diehl2015unsupervised} or only one trainable layer~\cite{masquelier2007unsupervised,beyeler2013categorization,kheradpisheh2016bio}.

In this paper we proposed a STDP-based spiking deep neural network (SDNN) with a spike-time neural coding. The network is comprised of a temporal-coding layer followed by a cascade of consecutive convolutional (feature extractor) and pooling layers. The first layer converts the input image into an asynchronous spike train, where the visual information is encoded in the temporal order of the spikes. Neurons in convolutional layers integrate input spikes, and emit a spike right after reaching their threshold. These layers are equipped with STDP to learn visual features. Pooling layers provide translation  invariance and also compact the visual information~\cite{Serre2007.PAMI}. Through the network, visual features get larger and more complex, where neurons in the last convolutional layer learn and detect object prototypes. At the end, a classifier detects the category of the input image based on the activity of neurons in the last pooling layer with global receptive fields.

We evaluated the proposed SDNN on Caltech face/motorbike and ETH-80 datasets with large-scale images of various objects taken form different viewpoints. The proposed SDNN reached the accuracies of 99.1\% on face/motorbike task and 82.8\% on ETH-80, which indicates its capability to recognize several natural objects even under severe variations. Based on our knowledge, there is no other spiking deep network which can recognize large-scale natural objects. We also examined the proposed SDNN on the MNIST dataset which is a benchmark for spiking neural networks, and interestingly, it reached 98.4\% recognition accuracy. In addition to the high performance, the proposed SDNN is highly energy-efficient and works with a few number of spikes per image, which makes it suitable for neuromorphic hardware implementation. Although current state-of-the-art DCNNs achieved stunning results on various recognition tasks, continued work on  brain-inspired models could end up in strong intelligent systems in future, which can even help us to improve our understanding of the brain itself.

\section*{Proposed Spiking Deep Neural Network}
\begin{figure*}[!h]
\centering
\includegraphics[scale=.95]{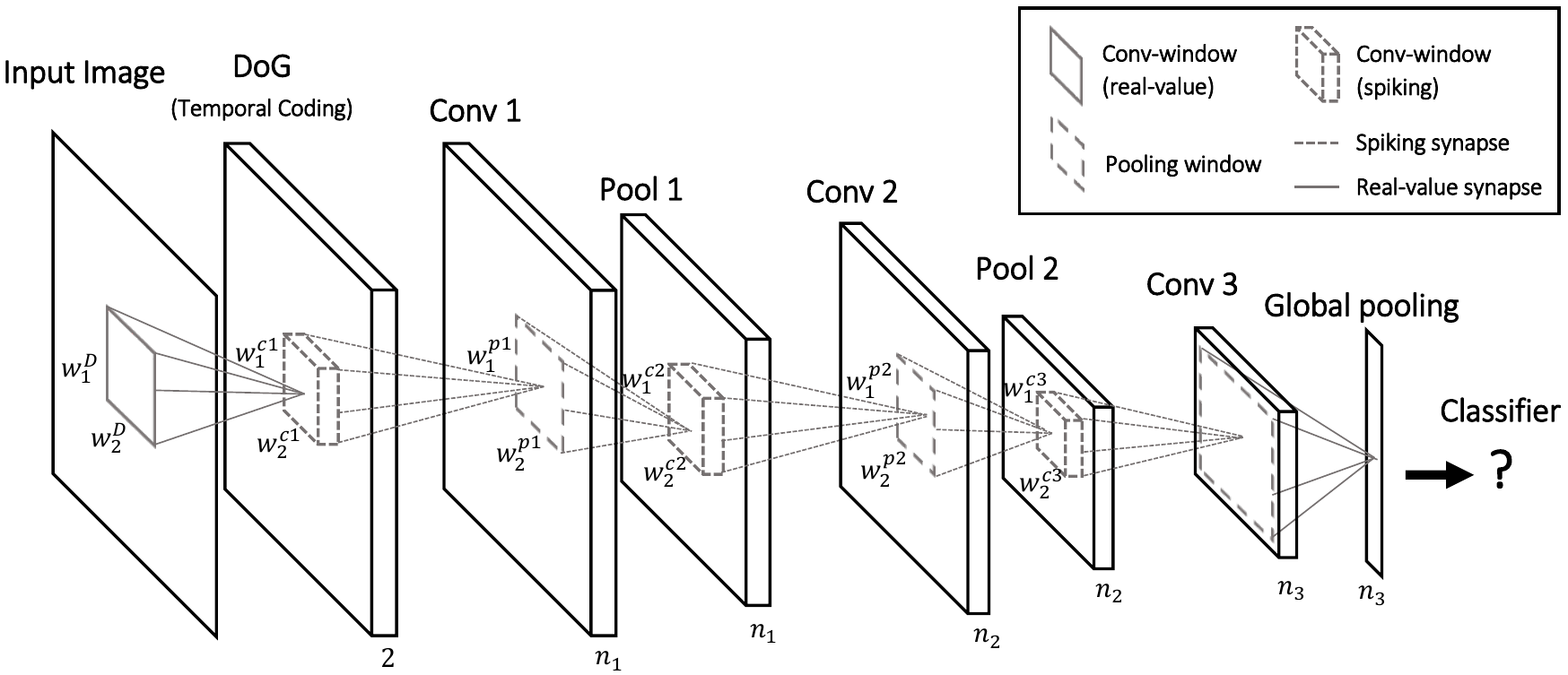}
\caption{ A sample architecture of the proposed SDNN with three convolutional and three pooling layers. The first layer applies ON- and OFF-center DoG filters of size $w_{1}^{D} \times w_{2}^{D}$  on the input image and encode the image contrasts in the timing of the output spikes. The $i$th convolutional layer, Conv $i$, learns combinations of features extracted in the previous layer. The $i$th pooling layer, Pool $i$, provides translation invariance for features extracted in the previous layer and compress the visual information using a local maximum operation. Finally the classifier detects the object category based on the feature values computed by the global pooling layer. The window size of the $i$th convolutional and pooling layers are indicated by $w_{1,2}^{ci}$ and $w_{1,2}^{pi}$, respectively. The number of the neuronal maps of the $i$th convolutional and pooling layer are also indicated by $n_{i}$ below each layer.}
\label{fig-network}
\end{figure*}

A sample architecture of the proposed SDNN with three convolutional and three pooling layers is shown in Fig.~\ref{fig-network}. Note that the architectural properties (e.g., the number of layers and receptive field sizes) and learning parameters should be optimized for the desired recognition task. 

The first layer of the network uses Difference of Gaussians (DoG) filters to detect contrasts in the input image. It encodes the strength of these contrasts in the latencies of its output spikes (the higher the contrast, the shorter the latency). Neurons in convolutional layers detect more complex features by integrating input spikes from the previous layer which detects simpler visual features. Convolutional neurons emit a spike as soon as they detect their preferred visual feature which depends on their input synaptic weights. Through the learning, neurons that fire earlier perform the STDP and prevent the others from firing via a winner-take-all mechanism. In this way, more salient and frequent features tend to be learned by the network.  Pooling layers provide translation invariance using maximum operation, and also help the network to compress the flow of visual data. Neurons in pooling layers propagate the first spike received from neighboring neurons in the previous layer which are selective to the same feature. Convolutional and pooling layers are arranged in a consecutive order. Receptive fields gradually increase through the network and neurons in higher layers become selective to complex objects or object parts. 

It should be noted that the internal potentials of all neurons are reset to zero before processing the next image. Also, learning only happens in convolutional layers and it is done layer by layer. Since the calculations of each neuron is independent of other adjacent neurons, to speed-up the computations, each of the convolution, pooling, and STDP operations are performed in parallel on GPU.

\subsection*{DoG and temporal coding}
The important role of the first stage in SNNs is to encode the input signal into discrete spike events in the temporal domain. This temporal coding  determines the content and the amount of information carried by each spike, which deeply affects the neural computations in the network. Hence, using efficient coding scheme in SNNs can lead to fast and accurate responses. Various temporal coding schemes can be used in visual processing (see ref.~\cite{thorpe2001spike}). Among them, rank-order coding is shown to be efficient for rapid processing (even possibly in retinal ganglion cells)~\cite{van2001rate,portelli2016rank}. 

Cells in the first layer of the network apply a DoG filter over their receptive fields to detect positive or negative contrasts in the input image. DoG well approximates the center-surround properties of the ganglion cells of the retina. When presented with an image, these DoG cells detect the contrasts and emit a spike; the more strongly a cell is activated (higher contrast), the earlier it fires.  In other word, the order of the spikes depends on the order of the contrasts. This rank-order coding is shown to be efficient for obtaining V1 like edge detectors~\cite{delorme2001networks} as well as complex visual features~\cite{masquelier2007unsupervised,kheradpisheh2016bio}  in higher cortical areas. 

DoG cells are retinotopically organized in two ON-center and OFF-center maps which are respectively sensitive to positive and negative contrasts. A DoG cell is allowed to fire if its activation is above a certain threshold. Note that this scheme grantees that at most one of the two cells (positive or negative) can fire in each location.  As mentioned above, the firing time of a DoG cell is inversely proportional to its activation value. In other words, if the output of the DoG filter at a certain location is $r$, the firing time of the corresponding cell is $t=1/r$. For efficient GPU-based parallel computing, the input spikes are grouped into equal-size sequential packets. At each time step, spikes of one packet are propagated simultaneously. In this way, a packet of spikes with near ranks (carrying similar visual information) are propagated in parallel, while, the next spike packet will be processed in the next time step.
\subsection*{Convolutional layers}
A convolutional layer contains several neuronal maps. Each neuron is selective to a visual feature determined by its input synaptic weights. Neurons in a specific map detect the same visual feature but at different locations. To this end, synaptic weights of neurons belonging to the same map should always be the same (i.e., weight sharing). Within a map, neurons are retinotopically arranged. Each neuron receives input spikes from the neurons located in a determined window in all neuronal maps of the previous layer. Hence, a visual feature in a convolutional layer is a combination of several simpler feature extracted in the previous layer. Note that the input windows of two adjacent neurons are highly overlapped. Hence, the network can detect the appearance of the visual features in any location.

Neurons in all convolutional layers are non-leaky integrate-and-fire neurons, which gather input spikes from presynaptic neurons and emit a spike when their internal potentials reach a prespecified threshold. Each presynaptic spike increases the neuron's potential by its synaptic weight. At each time step, the internal potential of the $i$th  neuron is updated as follows:
\begin{eqnarray}
\label{eq:neuron}
V_i(t)=V_i(t-1)+\sum_{j}{W_{j,i} S_j (t-1)},
\end{eqnarray}
where $V_i(t)$ is the internal potential of the $i$th convolutional neuron at time step $t$,  $W_{j,i}$ is the synaptic weight between the $j$th presynaptic neuron and the $i$th convolutional neuron, and $S_j$ is the spike train of the $j$th presynaptic neuron ($S_j(t-1)=1$ if the neuron has fired at time $t-1$, and $S_j(t-1)=0$ otherwise). If $V_i$ exceeds its threshold, $V_{thr}$, then the neuron emits a spike and $V_i$ is reset:
\begin{eqnarray}
\label{eq:neuron}
V_i(t)=0\:\: and \:\: S_i(t)=1, \quad if \:\: V_i(t)\geq V_{thr}.
\end{eqnarray}

Also, there is a lateral inhibition mechanism in all convolutional layers. When a neuron fires, in an specific location, it inhibits other neurons in that location belonging to other neuronal maps (i.e., resets their potentials to zero) and does not allow them to fire until the next image is shown. In addition, neurons are not allowed to fire more than once. These together provides an sparse but highly informative coding, because, there can be at most one spike at each location which indicates the existence of a particular visual feature in that location.

\subsection*{Local pooling layers}
Pooling layers help the network to gain invariance by doing a nonlinear max pooling operation over a set of neighboring neurons with the same preferred feature. Some evidence suggests that such a max operation occurs in complex cells in visual cortex~\cite{Serre2007.PAMI}. Thanks to the rank-order coding used in the proposed network, the maximum operation of pooling layers simply consists of propagating the first spike emitted by the afferents~\cite{rousselet2003taking}.

A neuron in a neuronal map of a pooling layer performs the maximum operation over a window in the corresponding neuronal map of the previous layer. Pooling neurons are integrate-and-fire neurons whose input synaptic weights and threshold are all set to one. Hence, the first input spike activates them and leads to an output spike. Regarding to the rank-order coding, each pooling neuron is allowed to fire at most once. It should be noted that no learning occurs in pooling layers. 

Another important role of pooling layers is to compress the visual information. Regarding to the maximum operation performed in pooling layers, adjacent neurons with overlapped inputs would carry redundant information (each spike is sent to many neighboring pooling neurons).  Hence, in the proposed network, the overlap between the input windows of two adjacent pooling neurons (belonging to the same map) is set to be very small. It helps to compress the visual information by eliminating the redundancies, and also, to reduce the size of subsequent layers.


\subsection*{STDP-based learning}
As mentioned above, learning occurs only in convolutional layers which should learn to detect visual features by combining simpler features extracted in the previous layer. The learning is done layer by layer, i.e., the learning in a convolutional layer starts when the learning in the previous convolutional layer is finalized. When a new image is presented, neurons of the convolutional layer compete with each other and those which fire earlier trigger STDP and learn the input pattern.

A simplified version of STDP~\cite{masquelier2007unsupervised} is used:
\begin{eqnarray}
\label{eq:STDP}
\Delta w_{ij}=
\left\{ 
  \begin{array}{l l l}
    a^{+}w_{ij}(1-w_{ij}),&  if & t_{j}-t_{i}\leq 0, \\
    a^{-}w_{ij}(1-w_{ij}),&  if & t_{j}-t_{i} > 0,
  \end{array} \right.
\end{eqnarray}
where $ i $ and $ j $ respectively refer to the index of post- and presynaptic neurons,
$t_i$ and $t_j$ are the corresponding spike times, $\Delta w_{ij}$ is the synaptic weight modification, and $a^{+}$ and $a^{-}$ are two parameters specifying the
learning rate. Note that the exact time difference between two spikes does not affect the weight change, but only its sign is considered. Also, it is assumed that if a presynaptic neuron does not fire before the postsynaptic one, it will fire later. These simplifications are equivalent to assuming that the intensity-latency conversion of DoG cells compresses the whole spike wave in a relatively short time interval (say, $20-30$ ms), so that all presynaptic spikes necessarily fall close to the postsynaptic spike time, and the time lags are negligible. The multiplicative term $w_{ij}(1-w_{ij})$ ensures the weights remain in the range [0,1] and thus maintains all synapses in an excitatory mode in adding to implementing soft-bound effect. 

Note that choosing large values for the learning parameters (i.e., $a^{+}$ and $a^{-}$) will decrease the learning memory, therefore, neurons would learn the last presented images and unlearn previously seen images. Also, choosing tiny values would slow down the learning process. At the beginning of the learning, when synaptic weights are random, neurons are not yet selective to any specific pattern and respond to many different patterns, therefore, the probability for a synapse to get depressed is higher than being potentiated. Hence, by setting $a^{-}$ to be greater than $a^{+}$, synaptic weights gradually decay insofar as neurons can not reach their threshold to fire anymore. Therefore, $a^{+}$ is better to be greater than $a^{-}$, however, by setting $a^{+}$ to be much greater than $a^{-}$, neurons will tend to learn more than one pattern and respond to all of them. All in all, it is better to choose $a^{+}$ and $a^{-}$ not too big and not too small, and it is better to set $a^{+}$ a bit greater than $a^{-}$.

During the learning of a convolutional layer, neurons in the same map, detecting the same feature in different locations, integrate input spikes and compete with each other to do the STDP. The first neuron which reaches the threshold and fires, if any, is the winner (global intra-map competition). The winner triggers the STDP and updates its synaptic weights. As mentioned before, neurons in different locations of the same map  have the same input synaptic weights (i.e., weight sharing) to be selective to the same feature. Hence, the winner neuron prevents other neurons in its own map to do STDP and duplicates its updated synaptic weights into them. Also, there is a local inter-map competition for STDP. When a neuron is allowed to do the STDP, it prevents the neurons in other maps within a small neighborhood around its location from doing STDP. This competition is crucial to encourage neurons of different maps to learn different features.

Because of the discretized time variable in the proposed model, it is probable that some competitor neurons fire at the same time step. One possible scenario is to pick one randomly and allow it to do STDP. But a better alternative is to pick the one which has the highest potential indicating higher similarity between its learned feature and input pattern.

Synaptic weights of convolutional neurons initiate with random values drown from a normal distribution with the mean of $\mu=0.8$ and STD of $\sigma=0.05$. Note that by choosing a small $ \mu $, neurons would not reach their threshold to fire and will not learn anything. Also, by  choosing a large $\sigma$, some initial synaptic weights will be smaller (larger) than others and have less (more) contribution to the neuron activity, and regarding the STDP rule, they have a higher tendency to converge to zero (one). In other words, dependency on the initial weights will be higher for a large $\sigma$.

 As the learning of a specific layer progresses, its neurons gradually converge to different visual features which are frequent in the input images. As mentioned before, learning in the subsequent convolutional layer statrs whenever the learning in the current convolutional layer is finalized. Here we measure the learning convergence of the $l$th convolutional layer as
\begin{equation}
C_l=\sum_{f}\sum_{i}{w_{f,i}(1-w_{f,i})}/n_w
\end{equation} 
where, $w_{f,i}$ is the $i$th synaptic weight of the $f$th feature and $n_w$ is the total number of synaptic weights (independent of the features) in that layer. $C_l$ tends to zero if each of the synaptic weights converge towards zero or one. Therefore, we stop the learning of the $l$th convolutional layer, whenever $C_l$ was sufficiently close to zero (i.e. $C_l< 0.01$).


\subsection*{Global pooling and classification}
The global pooling layer is only used in the classification phase. Neurons of the last layer perform a global max pooling over their corresponding neuronal maps in the last convolutional layer. Such a pooling operation provides a global translation invariance for prototypical features extracted in the last convolutional layer. Hence, there is only one output value for each feature, which indicates the presence of that feature in the input image. The output of the global pooling layer over the training images is used to train a linear SVM classifier. In the testing phase, the test object image is processed by the network and the output of the global pooling layer is fed to the classifier to determine its category. 

To compute the output of the global pooling layer, first, the threshold of neurons in the last convolutional layer were set to be infinite, and then, their final potentials  (after propagating the whole spike train generated by the input image) were measured. These final potentials can be seen as the number of early spikes in common between the current input and the stored prototypes in the last convolutional layer. Finally, the global pooling neurons compute the maximum potential at their corresponding neuronal maps, as their output value.

\section*{Results}
\subsection*{Caltech face/motorbike dataset}

We evaluated our SDNN on the face and motorbike categories of the Caltech 101 dataset available at http://www.vision.caltech.edu (see Fig.~\ref{figure2} for sample pictures). The training set contains 200 randomly selected images per category, and remaining images constitute the test set. The test images are not seen during the learning phase but used afterward to evaluate the performance on novel images. This standard cross-validation procedure allows measuring the system's ability to generalize, as opposed to learning the specific training examples.  All images were converted to grayscale values and rescaled to be 160 pixels in height (preserving the aspect ratio). In all the experiments, we used linear SVM classifiers with penalty parameter $C=1.0$ (optimized by a grid search in the range of $(0,10]$). 

\begin{figure*}[!htb]
\centering
\includegraphics[scale=.58]{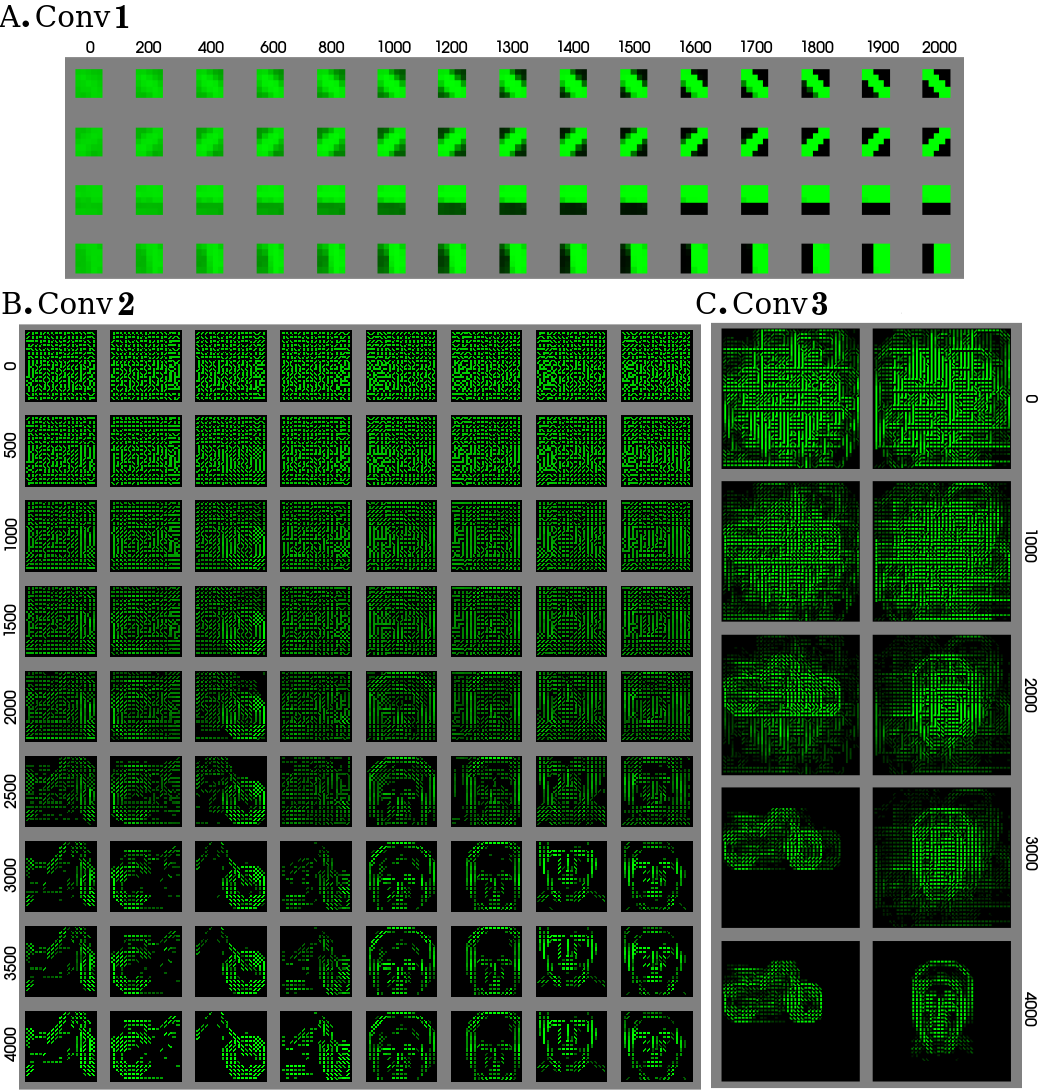}
\caption{ The synaptic changes of some neuronal maps in different layers through the learning with the Caltech face/motorbike dataset. A) The first convolutional layer becomes selective to oriented edges. B) The second convolutional layer converges to object parts. C) The third convolutional layer learns the object prototype and respond to whole objects.}
\label{figure1}
\end{figure*}

Here, we used a network similar to Fig.~\ref{fig-network}, with three convolutional layers each of which followed by a pooling layer.  For the first layer, only ON-center DoG filters of size $7 \times 7$ and standard deviations of 1 and 2 pixels are used. The first, second and third convolutional layers consists of 4, 20, and 10 neuronal maps with conv-window sizes of $5 \times 5$, $16 \times 16 \times 4$, and $5 \times 5 \times 20$ and firing thresholds of 10, 60, and 2, respectively. The pooling window sizes of the first and second pooling layers are $7 \times 7$ and $2 \times 2$ with the strides of 6 and 2, correspondingly. The third pooling layer performs a global max pooling operation. The learning rates of all convolutional layers are set to $a^{+}=0.004$ and $a^{-}=0.003$. In addition, each image is processed for 30 time steps.  

\begin{figure*}[!htb]
\centering
\includegraphics[scale=.2]{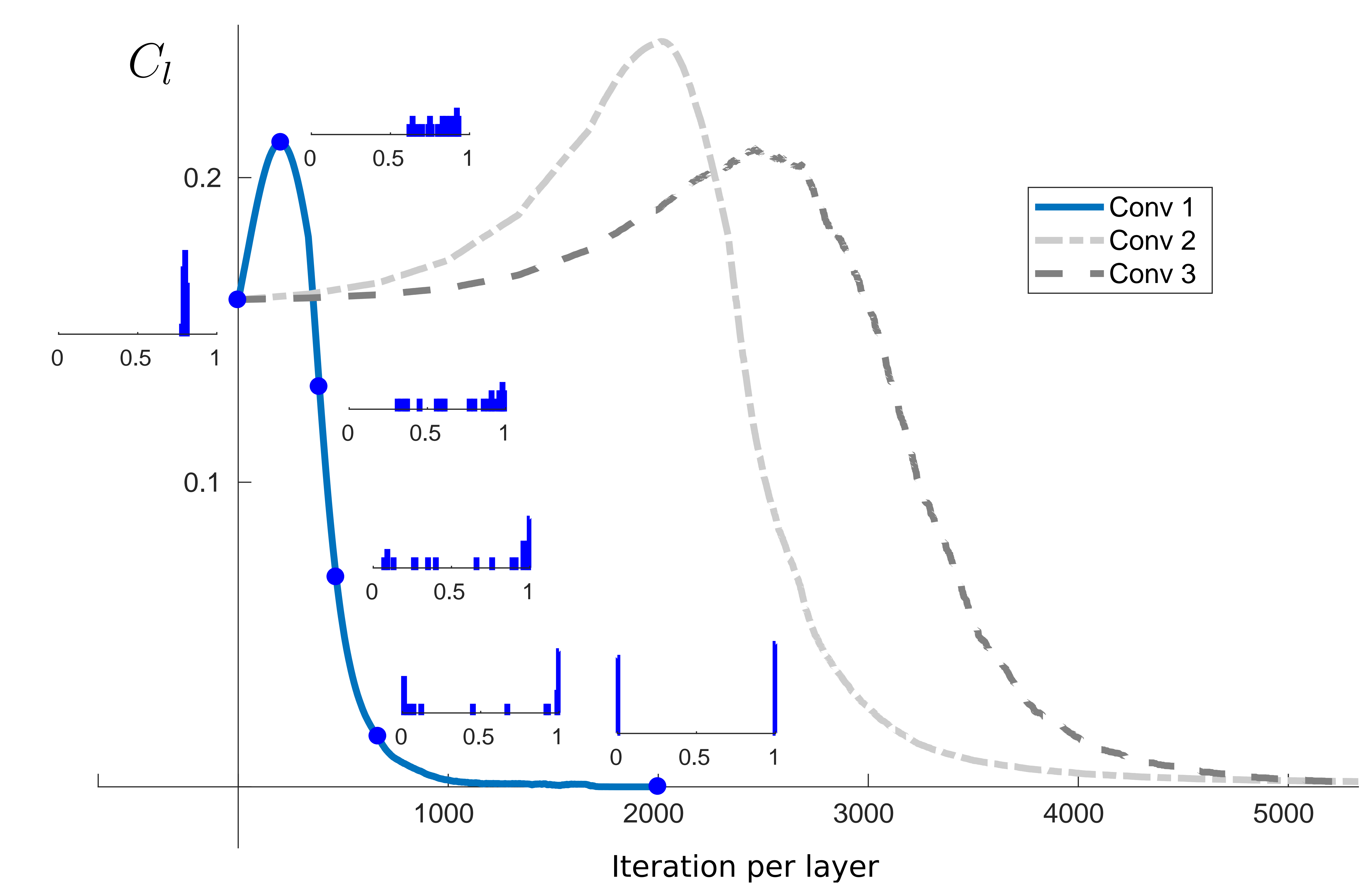}
\caption{Each curve shows the variation of the convergence index through the learning of a convolutonal layer. The weight histogram of the first convolutional layer at some critical points during the learning are shown next to its convergence curve.}
\label{figure2_2}
\end{figure*}

Fig.~\ref{figure1} shows the preferred visual features of some neuronal maps in the first, second and third convolutional layers through the learning process. To visualize the visual feature learned by a neuron, a backward reconstruction technique is used. Indeed, the visual features in the current layer can be reconstructed as the weighted combinations of the visual features in the previous layer. This backward process continues until the first layer, whose preferred visual features are computed by DoG functions. As shown in Fig.~\ref{figure1}A, interestingly, each of the four neuronal maps of the first convolutional layer converges to one of the four orientations: $\pi/4$, $\pi/2$, $3\pi/4$, and $\pi$. This shows how efficiently the association of the proposed temporal coding in DoG cells and unsupervised learning method (the STDP and learning competition) led to highly diverse edge detectors which can represent the input image with edges in different orientations. These edge detectors are similar to the simple cells in primary visual cortex (i.e., V1 area)~\cite{delorme2001networks}.

\begin{figure*}[!htb]
\centering
\includegraphics[scale=.55]{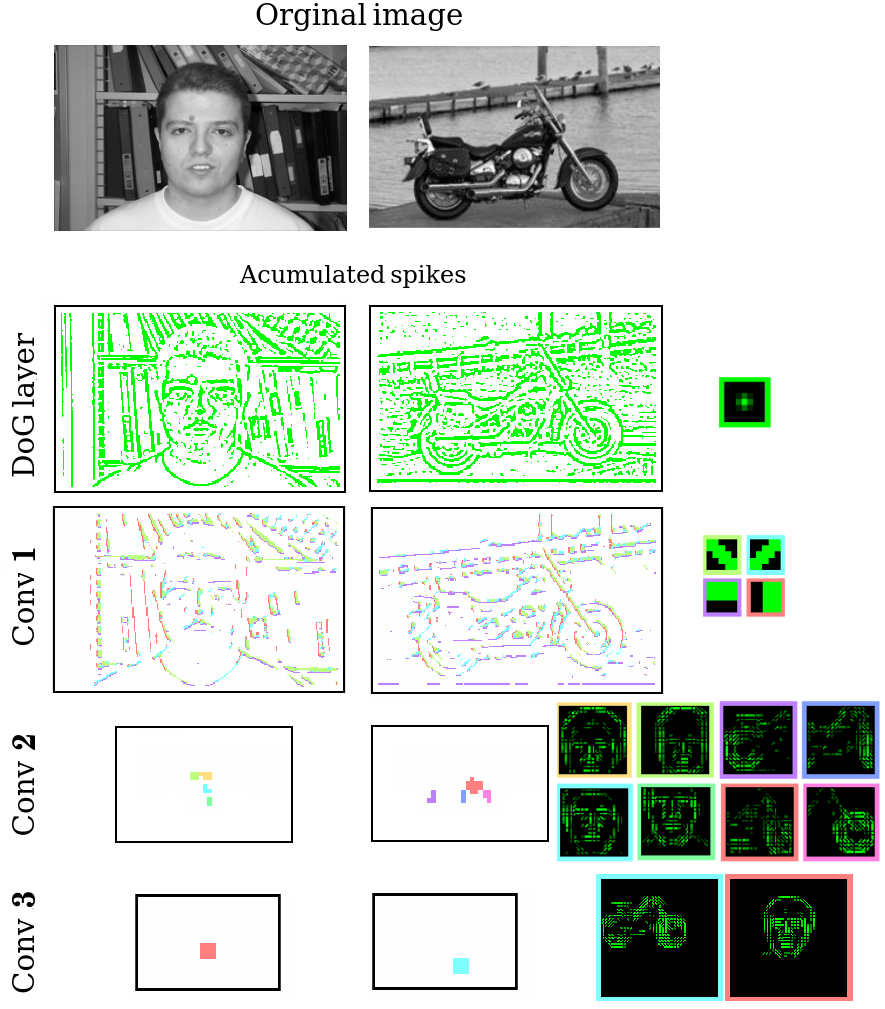}
\caption{ The spiking activity of the convolutional layers with the face and motorbike images. The preferred features of neuronal maps in each convolutional layer are shown on the right. Each feature is coded by a specific color border. The spiking activity of the convolutional layers, accumulated over all the time steps, is shown in the corresponding panels. Each point in a panel indicates that a neuron in that location has fired at a time step, and the color of the point indicates the preferred feature of the activated neuron.}
\label{figure2}
\end{figure*}

Fig.~\ref{figure1}B shows the learning progress for the neuronal  maps of the second convolutional layer. As mentioned, the first convolutional layer detects edges with different orientations all over the image, and due to the used temporal coding, neurons corresponding to edges with higher contrasts (i.e., salient edges) will fire earlier. On the other hand, STDP naturally tends to learn those combination of edges that are consistently repeating in the training images (i.e., common features between the target objects). Besides, the learning competition tends to prevent the neuronal maps from learning similar visual features. Consequently, neurons in the second convolutional layer learn the most salient, common, and diverse visual features of the target objects, and do not learn the backgrounds that drastically change between images. As seen in Fig.~\ref{figure1}B, each of the maps gradually learns a different visual feature (combination of oriented edges) representing a face or motorbike feature.

The learning progress for two neuronal maps of the third convolutional layer are shown in Fig.~\ref{figure1}C.  As seen, one of them gradually becomes selective to a complete motorbike prototype as a combination of motorbike features such as back wheel, middle body, handle, and front wheel detected in the second layer. Also, the other map learns a whole face prototype as a combination of facial features. Indeed, the third convolutional layer learns the whole object prototypes using intermediate complexity features detected in the previous layer. Neurons in the second layer compete with each other and send spikes toward the third layer as they detect their preferred visual features. Since, different combinations of these features are detected for each object category, neuronal maps of the third layer will learn different prototypes of different categories. Therefore, the STDP and the learning competition mechanism direct neuronal maps of the third convolutional layer to learn highly category specific prototypes.

As mentioned in the previous section, learning at the $l$th convolutional layer stops whenever the learning convergence index $C_l$ tends to zero. Here, in Fig.~\ref{figure2_2}, we presented the evolution of $C_l$ during the learning of each convolutional layer. Also, we provided the weight histogram of the first convolutional layer at some critical points during the learning process. Since the initial synaptic weights are drown from a  normal distribution with $\mu=0.8$ and $\sigma=0.05$, at the beginning of the learning, the convergence index of the  $l$th layer starts from $C_l\simeq 0.16$. Since features are not formed at the early iterations, neurons respond to  almost every pattern, therefore many of the synapses are depressed most of the times and gradually move towards 0.5, and $C_l$ peaks. As a consequence, neurons' activity  decreases, and they start responding to a few of the patterns and not to others. From then, synapses contributed in these patterns are repeatedly potentiated and others are depressed. Due to the nature of the employed soft-bound STDP, learning is faster when the weights are around 0.5 and $C_l$ rapidly decreases. Finally, at the end of the learning, as features are formed and synaptic weights converge to zero or one, $C_l$ tends to zero. 

As seen in  Fig.~\ref{figure2_2}, the weight changes in lower layers are faster than in higher layers. This is mainly due to the stronger competition among the feature maps in higher layers. For instance, in the first layer, there are only four feature maps with small receptive fields. If a neuron from one feature map loses the competition at one location, another neuron can still win the competition at some other location and do the STDP. While, in the third layer, there are more feature maps with larger receptive fields. If a neuron wins the competition, it prevents neurons of other feature maps from doing the STDP in a much wider area. Hence, for each image, only a few of the feature maps update their weights.  Therefore, the competition is stronger in higher layers and learning takes a much longer time. Beside the competition factor, neurons in higher layers become selective to more complex and less frequent features so they do not reach their threshold as often as  neurons  in lower layers (e.g., all images contain edges).

Fig.~\ref{figure2} shows the accumulated spiking activity of the DoG and the following three convolutional layers over all time steps, for two face and motorbike sample images. For each layer, the preferred features of some neuronal maps with color coded borders are demonstrated on top, and their corresponding spiking activity are shown in  panels below them. Each colored point inside a panel indicates the neuronal map of the neuron which has fired in that location at a time step. As seen, neurons of the DoG layer detect image contrasts, and edge detectors in the first convolutional layer detect the orientation of edges. Neurons in the second convolutional layer, which are selective to intermediate complexity features, detect their preferred visual feature by combining input spikes from edge detector cells in the first layer. Finally, the coincidence of these features activates neurons in the third convolutional layer which are selective to object prototypes.  As seen, when a face (motorbike) image is presented, neurons in the face (motorbike) maps fire. To better illustrate the learning progress of all the layers as well as their spiking activity in the temporal domain, we prepared a short video (see \nameref{S1_Video}). 

\begin{figure}[!tb]
\centering
\includegraphics[scale=.51]{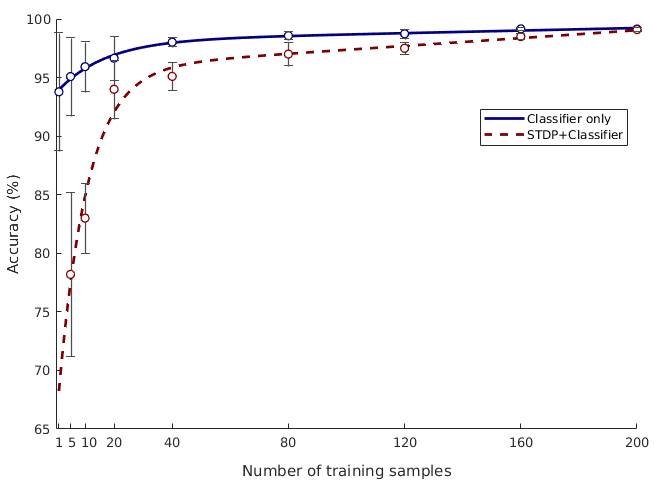}
\caption{ Recognition accuracies (mean $\pm$ std) of the proposed SDNN for different number of training images per category used in  the STDP-based feature learning and/or training the classifier. The red curve presents the model's accuracy when the different number of images are used to train the network (by unsupervised STDP) and the classifier as well. The blue curve shows the model's accuracy when the layers of the network are trained using STDP with 400 images (200 from each category), and the classifier is trained with different number of labeled images per category.}
\label{figure2_5}
\end{figure}

As mentioned in the previous section, the output of the global pooling layer is used by a linear SVM classifier to specify the object category of the input images. We trained the proposed SDNN on training images and evaluated it over the test images, where the model reached the categorization accuracy of $99.1\pm 0.2\%$. It shows how the object prototypes, learned in the highest layer, can well represent the object categories. Furthermore, we also calculated the  single neuron accuracy. In more details, we separately computed the recognition accuracy of each neuron in the global pooling layer. Surprisingly, some single neurons reached an accuracy of $93\%$,  and the mean accuracy was 89.8\%. Hence, it can be said that single neurons in the highest layer are highly class specific, and different neurons carry complementary information which altogether provide robust object representations.

\begin{table*}[!htb]
\centering
\caption{Recognition accuracies of the proposed SDNN with random features in different convolutional layers.}
\label{tab0}
\begin{tabular}{|c|c|c|c|c|}
\hline
Conv layers with random features &  Non& 3rd& 2nd \& 3rd & 1st \& 2nd \& 3rd\\ \hline
Accuracy (\%) &99.1&80.2&67.8&66.3\\
\hline
\end{tabular}
\end{table*}

\begin{table*}[!htb]
\centering
\caption{Recognition accuracies of the proposed SDNN for differen amounts of noise.}
\label{tab00}
\begin{tabular}{|c|c|c|c|c|c|c|c|}
\hline
Noise level &  $\alpha=0\%$&  $\alpha=5\%$ & $\alpha=10\%$& $\alpha=20\%$ & $\alpha=30\%$& $\alpha=40\%$ & $\alpha=50\%$\\ \hline
Accuracy (\%) &99.1&95.4&91.6&84.3&63.7&57.6&54.2\\
\hline
\end{tabular}
\end{table*}

To better demonstrate the role of the STDP learning rule in the proposed SDNN, we used random features in different convolutional layers and assessed the final accuracy. To this end, we first trained all the three convolutional layers of the network (using STDP) until they all converged (synaptic weights had a bimodal distribution with 0 and 1 as centers). Then, for a certain convolutional layer, we counted the number of active synapses (i.e., close to one) of each of its feature maps. Corresponding to each learned feature in the first convolutional layer, we generated a random feature with the same number of active synapses. For the second and third convolutional layers, the number of active synapses in the random features was doubled. Note that with fewer active synapses, neurons in the second and third convolutional layers could not reach their threshold, and with more active synapses, many of the neurons tend to fire together. Anyways, we evaluated the network using these random features. Table~\ref{tab0} presents the SDNN accuracy when we had random features in the third, or second and third, or in all the three convolutional layers. As seen, by replacing the learned features of the lower layers with random ones, the accuracy decreases more. Especially,  using random features in the second layer, the accuracy severely drops. This shows the critical role of intermediate complexity features for object categorization.

We also evaluated how the proposed SDNN is robust to noise.  To this end, we added a white noise to the neurons' threshold during both training and testing phases. Indeed, for each image, we added to the threshold of each neuron, $V_{thr}$, a random value drawn from a uniform distribution in range $\pm \alpha\%$ of $V_{thr}$. We evaluated the proposed DCNN for different amount of noise (from  $\alpha=5\%$ to $50\%$) and the results are provided in Table~\ref{tab00}. Up to the 20\% of noise, the accuracy is still reasonable, but by increasing the noise level, the accuracy dramatically drops and reaches to the chance level in case of 50\% of noise. In other words, the network is more or less able to tolerate the instability caused by the noise below 20\%, but when it goes further, the neurons' behavior drastically change during the learning  and STDP can not extract informatic features from the input images.

\begin{figure*}[!tb]
\centering
\includegraphics[scale=.46]{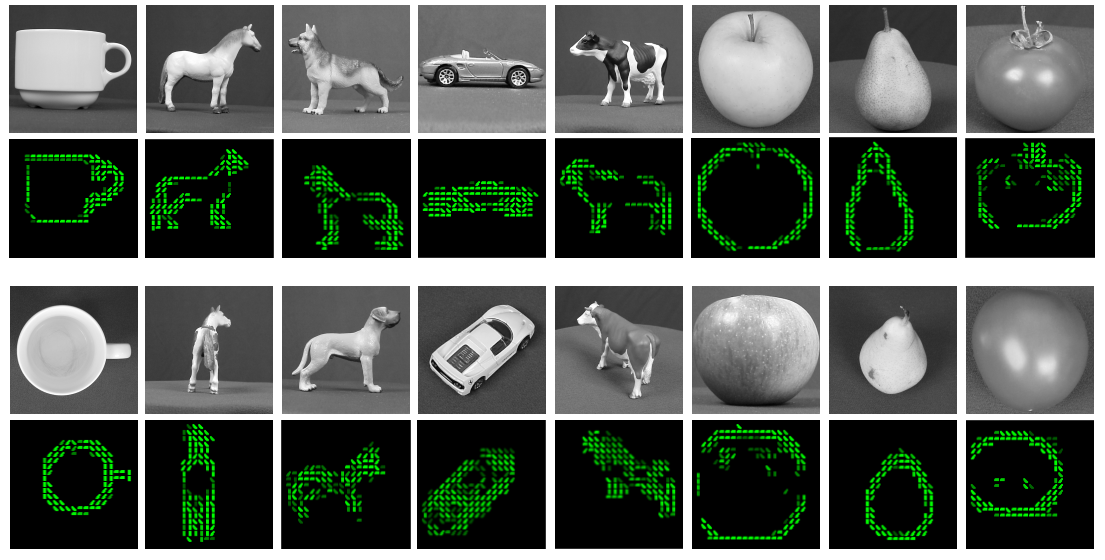}
\caption{ Some sample images of different object categories of ETH-80 in different viewpoints. For each image, the preferred feature of an activated neuron in the third convolutional layer is shown in below.}
\label{figure3}
\end{figure*}

In another experiment, we changed the number of training samples and calculated the recognition accuracy of the proposed SDNN. For instance, we trained the network and the classifier with 5 samples from each category, and then evaluated the final system with the test samples. As shown by the red curve in Fig.~\ref{figure2_5}, with 5 images per category the model reached the accuracy of 78.2\%, and only 40 images from each category are sufficient to reach 95.1\% recognition accuracy. Although having more training samples leads to higher accuracies, the proposed SDNN can extract diagnostic features and reach reasonable accuracies even using a few tens of training images.  Due to the unsupervised nature of STDP, the proposed SDNN does not suffer much from the overfitting challenge caused by small training set size in supervised learning algorithms such as back-propagation. In a real world, the number of labeled samples is very low. Therefore, learning in humans or other primates is mainly unsupervised. Here, in another experiment, we trained the network using STDP over 200 images per category, then we used different portions of these images as labeled samples (from 1 image to 200 images per category) to train the classifier. It lets us see whether the visual feature that the network has learned from the unlabeled images (using unsupervised STDP)  is sufficient for solving the categorization task with only a few labeled training samples. As shown by the blue curve in Fig.~\ref{figure2_5}, with only one sample per category the model could reach the average accuracy of 93.8\%. It suggests that the unsupervised STDP can provide a rich feature space well explaining the object space and reducing the need for labeled examples.

\subsection*{ETH-80 dataset}
The ETH-80 dataset contains eight different object categories: apple, car, toy cow, cup, toy dog, toy horse, pear, and tomato (10 instances per category). Each object is photographed from 41 view- points with different view angles and different tilts. Some examples of objects in this dataset are shown in Fig.~\ref{figure3}. ETH-80 is a good benchmark to show how the proposed SDNN can handle multi-object categorization tasks with high inter-instance variability, and how it can tolerate huge view-point variations. In all the experiments, we used linear SVM classifiers with penalty parameter $C=1.2$ (optimized by a grid search in the range of $(0,10]$).

Five randomly chosen instances of each object category are selected for the training set  used in the learning phase. The remaining instances constitute the testing set, and are not seen during the learning phase. All the object images were converted to grayscale values. To evaluate the proposed SDNN on ETH-80, we used a network architecturally similar to the one used for Caltech face/motorbike dataset. The other parameters are also similar, except for the number of neuronal maps in the second and third convolutional and pooling layers. Here we used 400 neuronal maps in each of these layers.

\begin{table}[!t]
\caption{Recognition accuracies of the proposed SDNN and some other methods over the ETH-80 dataset. Note that all the models are trained on 5 object instances of each category and tested on other 5 instances.}
\label{tab1}
\centering
\begin{tabular}{|c|c|}
\hline
Method & Accuracy (\%)\\ \hline
HMAX~\cite{kheradpisheh2016bio} & 69.0\\ \hline
Convolutional SNN~\cite{kheradpisheh2016bio} & 81.1\\ \hline
Pre-trained AlexNet &  79.5\\ \hline
Fine-tuned AlexNet &  94.2\\ \hline
Supervised DCNN &  81.9\\ \hline
Unsupervised DCA &  80.7\\ \hline
Proposed SDNN & 82.8\\
\hline
\end{tabular}
\end{table}

Similar to the caltech dataset, neuronal maps of the first convolutional layer converged to the four oriented edges. Neurons in the second and third convolutional layers also became selective to intermediate features and object prototypes, respectively. Fig.~\ref{figure3}  shows sample images from the ETH-80 dataset and the preferred features of some neuronal maps in the  third convolutional layer which are activated for those images. As seen, neurons in the highest layer respond to different views of different objects, and altogether, provide an invariant object representation. Thus, the network learns  2D and view-dependent prototypes of each object category to achieve 3D representations.

As mentioned before, we evaluated the proposed SDNN over the test instances of each object category which are not shown to the network during the training. The recognition accuracy of the proposed SDNN along with some other models on the ETH-80 dataset are presented in Table~\ref{tab1}. HMAX~\cite{serre2007feedforward} is one of the classic computational models of the object recognition process in visual cortex. It has 5000 features and uses a linear SVM as the classifier (see~\cite{kheradpisheh2016bio} for more details). Convolutional SNN~\cite{kheradpisheh2016bio} is an extension of the Masquelier et al. 2007~\cite{masquelier2007unsupervised} which had  one trainable layer with 1200 visual features and used linear SVM for classification. AlexNet is the first DCNN that significantly improved recognition accuracy on Imagenet Dataset. Here, we compared our proposed SDNN with both Imagenet pre-trained and ETH-80 fine-tuned versions of AlexNet. To obtain the accuracy of the pre-trained AlexNet, images were shown to the model and feature vectors of the last layer were used to train and test a linear SVM classifier. Also, to fine-tune the Imagenet pre-trained AlexNet on ETH-80, its decision layer was replaced by an eight-neuron decision layer and trained by the stochastic gradient descent learning algorithm.

Regarding the fact that the pre-trained AlexNet has not seen ETH-80 images and fine-tuned AlexNet has already trained by millions of images from Imagenet dataset, the comparison may not be fair enough. Therefore, we compared the proposed SDNN to a \textit{supervised DCNN} with the same structure but having two extra dense and decision layers on top with 70 and 8 neurons, respectively. The DCNN was trained on the gray-scaled images of ETH-80 using the backpropagation algorithm. The ReLU and soft-max activation functions were employed for the intermediate and decision layers, respectively. We used a cross-entropy loss function with L2 kernel regularization and L1 activity regularization terms. We also performed a 50\% dropout regularization on the dense layer. The hyper-parameters such as learning rates, momentums, and regularization factors were optimized using a grid search. Also, we used an \textit{early stopping startegy} to prevent the network from overfitting. Eventually, the supervised DCNN reached the average accuracy of $81.9\%$ (see Table~\ref{tab1}). Note that we tried to evaluate the DCNN with a dense layer of more than 70 neurons, but all the time the network got quickly  overfitted on the training data with no accuracy improvement over the test set. It seems that the supervised DCNN suffers from the lack of sufficient training data.

In addition, we compared the proposed SDNN to a deep convolutional autoencoder (DCA) which is one of the best unsupervised learning algorithms in machine learning. We developed a DCA with an encoder network having the same architecture as our SDNN followed by a decoder network with reversed architecture. The ReLU activation function was used in the convolutional layers of both encoder and decoder networks. We used the cross-entropy loss function and stochastic gradient descent learning algorithm to train the DCA. The learning parameters (i.e., learning rate and momentum) were optimized through a grid search. When the learning had converged, we eliminated the decoder part and used the encoder's output representations to train and test a linear SVM classifier. Note that DCA was trained on gray-scaled images of ETH-80. The DCA reached the average accuracy of 80.7\% (see Table~\ref{tab1}) and the proposed SDNN could outperform the DCA network with the same structure. 


We also evaluated the proposed SDNN that has two convolutional layers. Indeed, we removed the third convolutional layer, applied the global pooling over the second convolutional layer, and trained a new SVM classifier over its output. The model's accuracy dropped by 5\% and reached to $77.4\%$. Although the features in the second layer represent object parts with intermediate complexity, it is the combination of these features in the last layer which culminates the object representations and makes the classification easier. In other words, the  similarity between the smaller parts of objects of differen categories but with similar shapes (e.g., apple and potato) is higher than the whole objects, hence, the visual features in the higher layers can provide better object representations.

\begin{figure}[!t]
\centering
\includegraphics[scale=.59]{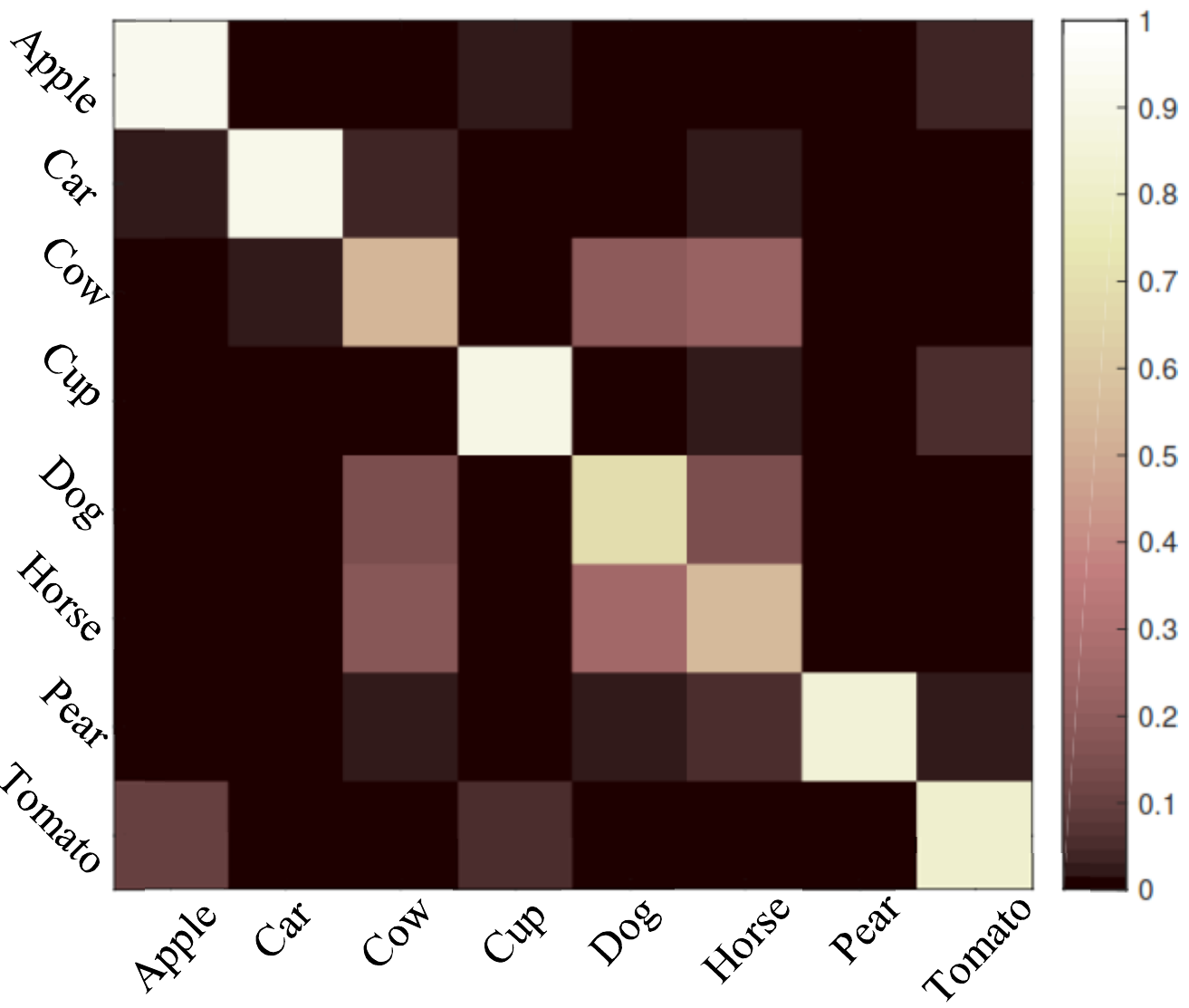}
\caption{ The confusion matrix of the proposed SDNN over the ETH-80 dataset.}
\label{figure4}
\end{figure}

\begin{figure}[!t]
\centering
\includegraphics[scale=.45]{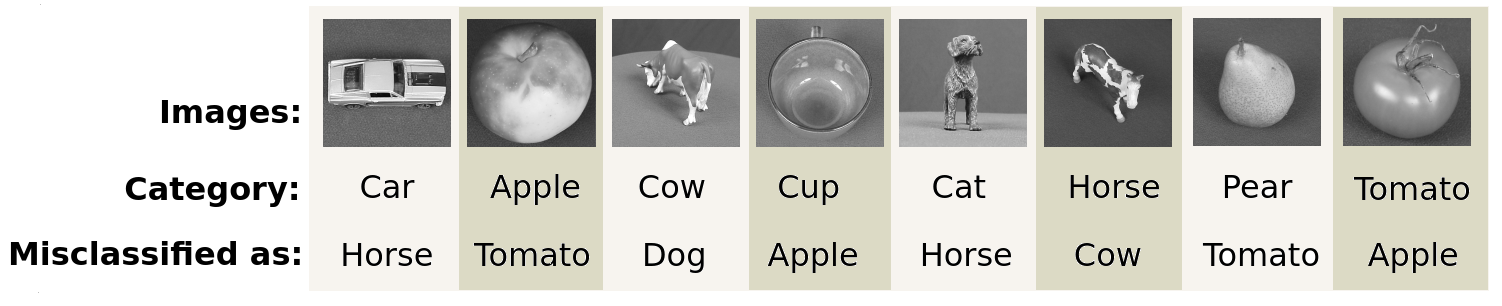}
\caption{ Some misclassified samples of the ETH-80 dataset by the proposed SDNN.}
\label{figure4_2}
\end{figure}

In a subsequent analysis, we computed the confusion matrix, to see which categories are mostly confused with each other. Fig.~\ref{figure4} illustrates the confusion matrix of the proposed SDNN over the ETH-80 dataset. As seen, most of the errors are due to the miscategorization of dogs, horses, and cows. We checked whether these errors belong to the some specific viewpoints or not. We found out that the errors are uniformly distributed between different viewpoints. Some misclassified samples are shown in Fig.~\ref{figure4_2}. Overall, it can be concluded that these categorization errors are due to the overall shape similarity between these object categories. 

The other important aspect of the proposed SDNN is the computational efficiency of the network. For each ETH-80 image, on average, about 9100 spikes are emitted in all the layers, i.e., about 0.02 spike per neuron per image. Note that the number of inhibitory events is equal to the number of spikes. These together points to the fact that the proposed SDNN can recognize objects with high precision but low computational cost. This efficiency is caused by the association of the proposed temporal coding and STDP learning rule which led to a sparse but informative visual coding. 

\subsection*{MNIST dataset}
MNIST~\cite{lecun1998gradient} is a benchmark dataset for SNNs which has been widely used~\cite{hussain2014improved,zhao2015feedforward,querlioz2013immunity,o2015real,diehl2015unsupervised,diehl2015fast}. We also evaluated our SDNN on the MNIST dataset which contains 60,000 training and 10,000 test handwritten single-digit images.  Each image is of size $28 \times 28$ pixels and contains one of the digits 0--9. For the first layer, ON- and OFF-center DoG filters with standard deviations of 1 and 2 pixels are used. The first and second convolutional layers respectively consist of 30 and 100 neuronal maps with $5 \times 5$ convolution-window and firing thresholds of 15 and 10. The pooling-window of the first pooling layer was of size $2 \times 2$ with the stride of 2. The second pooling layer performs a global max operation. Note that the learning rates of all convolutional layers were set to $a^{+}=0.004$ and $a^{-}=0.003$. In all the experiments, we used linear SVM classifiers with penalty parameter $C=2.4$ (optimized by a grid search in the range of $(0,10]$).

\begin{figure}[!tb]
\centering
\includegraphics[scale=.6]{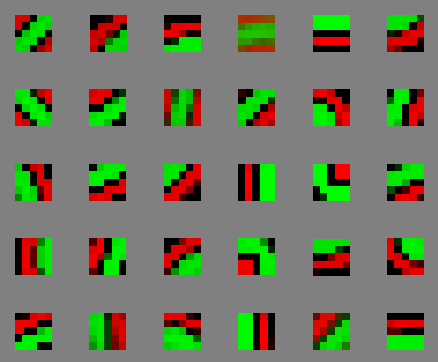}
\caption{The Gabor-like features learned by the neuronal maps of the first convolutional layer from the MNIST images. The red and green colors receptively indicate the  strength of input synapses from ON- and OFF-center DoG cells.}
\label{figure5}
\end{figure}

Fig.~\ref{figure5} shows the preferred features of some neuronal maps in the first  convolutional layer. The green and red colors correspond to ON- and OFF-center DoG filters. Interestingly, this layer converged to Gabor-like edge detectors with different orientations, phase and polarity. These edge features are combined in the next layer and provide easily separable digit representation.

 \begin{table*}[!ht]
 \centering
\caption{Recognition accuracies of the proposed SDNN and some other SNNs over the MNIST dataset.}
\label{tab2}
{\begin{tabular}{|c|c|c|c|c|}
\hline
Architecture &Neural coding& Learning-type & Learning-rule & Accuracy (\%)\\ \hline
Dendritic neurons~\cite{hussain2014improved} & Rate-based& Supervised & Morphology learning & 90.3\\
\hline
Convolutional SNN~\cite{zhao2015feedforward} & Spike-based & Supervised & Tempotron rule & 91.3\\
\hline
Two layer network~\cite{querlioz2013immunity} & Spike-based &Unsupervised &  STDP & 93.5\\
\hline
Spiking RBM~\cite{o2015real}& Rate-based & Supervised & Contrastive divergence & 94.1\\
\hline
Two layer network~\cite{diehl2015unsupervised} & Spike-based & Unsupervised &  STDP &
95.0\\
\hline
Convolutional SNN~\cite{diehl2015fast} & Rate-based & Supervised & Back-propagation & \textbf{99.1}\\
\hline
Proposed SDNN & Spike-based & Unsupervised & STDP & 98.4\\
\hline
\end{tabular}}
\end{table*}

  Recognition performance of the proposed method and some recent SNNs on the MNIST dataset are provided in Table~\ref{tab2}. As seen, the proposed SDNN outperforms unsupervised SNNs by reaching $98.4\%$ recognition accuracy. Besides, the accuracy of the proposed SDNN is close to the $99.1\%$ accuracy of the totally supervised rate-based SDNN~\cite{diehl2015fast} which is indeed the converted version of a traditional DCNN trained by back-propagation. 
 
The important advantage of the proposed SDNN is the use of much fewer spikes. Our SDNN  uses only about 600 spikes for each MNIST images in total for  all the layers, while the supervised rate-based SDNN uses thousands of spikes per layer~\cite{diehl2015fast}. Also, because of using rate-based neural coding in such networks, they need to process images for hundreds of time steps, while our network process the MNIST images in 30 time steps only. Notably, whenever a neuron in our network fires it inhibits other neurons at the same position, therefore, the total number of inhibitory events per each MNIST image is equal to the number of spikes(i.e., 600). As stated in Section 2, the proposed SDNN uses a temporal code which encodes the information of the input image in the spike times, and each neuron in all  layers, is allowed to fire at most once. This temporal code associated with unsupervised STDP rule leads to a fast, accurate, and efficient processing.

\section*{Discussion}
Recent supervised DCNNs have reached high accuracies on the most challenging object recognition datasets such as Imagenet. Architecture of theses networks are largely inspired by the deep hierarchical processing in the visual cortex. For instance, DCNNs  use retinotopically arranged neurons with restricted receptive fields, and the receptive field size and feature complexity of the neurons gradually increase through the layers. However, the learning and neural processing mechanisms applied in DCNNs are inconsistent with the visual cortex, where neurons communicate using spikes and learn the input spike patterns in a mainly unsupervised manner. Employing such mechanisms in DCNNs can improve their energy consumption and decrease their need for an expensive supervised learning with millions of labeled images.

A popular approach in previous researches is to convert pre-trained supervised DCNNs into equivalent spiking network. To simulate the floating-point calculations in DCNNs, they  have to use the firing rate as the neural code, which in result increases the number of required spikes and the processing time. For instance, the converted version of a simple two-layer DCNN for the MNIST dataset with images of $28 \times 28$ pixels requires thousands of spikes and hundreds of time steps per image. On the other hand, there are some SDNNs \cite{panda2016unsupervised,burbank2015mirrored,bengio2015towards} which are originally spiking network but employ firing rate coding and  biologically implausible learning rules such as autoencoders and back-propagation.

Here, we proposed a STDP-based SDNN with a spike-time coding. Each neuron was allowed to fire at most once, where its spike-time indicates the significance of its visual input. Therefore, neurons that fire earlier are carrying more salient visual information, and hence, they were allowed to do the STDP and learn the input patterns. As the learning progresses, each layer converged to a set of diverse but informative features, and the feature complexity gradually increases through the layers from simple edge features to object prototypes. The proposed SDNN was evaluated on several image datasets and reached high recognition accuracies. This shows how the proposed temporal coding and learning mechanism (STDP and learning competition) lead to  discriminative object representations. 

The proposed SDNN has several advantages to its counterparts. First, our proposed SDNN is the first spiking neural network with more than one learnable layer which can process large-scale natural object images. Second, due to the use of an efficient temporal coding, which encodes the visual information in the time of the first spikes, it can process the input images with a low number of spikes and in a few processing time steps. Third, the proposed SDNN exploits the bio-inspired and totally unsupervised STDP learning rule which can learn the diagnostic object features and neglect the irrelevant backgrounds.

We compared the proposed SDNN to several other networks including unsupervised methods such as HMAX and convolutional autoencoder network, and supervised methods such as DCNNs. The proposed SDNN could outperform the unsupervised methodes which shows its advantages in extracting more informatic features from training images. Also, it was better than the supervised deep network which largely suffered from overfitting and lack of sufficient training data. Although, the state-of-the-art supervised DCNNs have stunning performance on large datasets like Imagenet, but contrary to the proposed SDNN, they fall in trouble with small and mideum size datasets.

Our SDNN could be efficiently implemented in parallel hardware (e.g., FPGA~\cite{Yousefzadeh2015}) using address event representation (AER)~\cite{Sivilotti1991} protocol. With AER, spike events are represented by the addresses of sending and receiving neurons, and time is represented by the asynchronous occurrence of spike events. Since these hardware are much faster than biological hardware, simulations could run several order of magnitude faster than real time~\cite{Serrano-Gotarredona2013}. The primate visual system extracts the rough content of an image in about 100 ms~\cite{thorpe1996speed,hung2005fast,Kirchner2006,liu2009timing}. We thus speculate that some dedicated hardware will be able to do the same in the order of a millisecond or less.

Also, the proposed SDNN can be modified to use spiking retinal models~\cite{Wohrer2009,Martinez-Canada2016} as the input layer. These models mimic the spatiotemporal filtering  of the retinal ganglion cells with center/surround receptive fields. Alternatively, we could use neuromorphic asynchronous event-based cameras such as dynamic vision sensor (DVS), which generate output events when they capture transients in the scene~\cite{Lichtsteiner2007}. Finally, due to the DoG filtering in the input layer of the proposed SDNN, some visual information such as texture and color are lost. Hence, future studies should focus on encoding these additional pieces of information in the input layer.

Biological evidence indicate that in addition to the unsupervised learning mechanisms (e.g., STDP), there are also dopamine-based reinforcement learning strategies in the brain~\cite{pignatelli2015role}. Besides, although it is still unclear how supervised learning is implemented in biological neural networks,  it seems that for some tasks (e.g., motor control and sensory inputs prediction) the brain must constantly learn temporal dynamics based on error feedback~\cite{doya2000complementary}. Employing such reinforcement and supervised learning strategies could improve the proposed SDNN in different aspects which are inevitable with unsupervised learning methods. Particularly, they can help to reduce the number of required features and to extract optimized task-dependent features.

\section*{Supporting Information}


\paragraph*{Video 1}
\label{S1_Video}
We prepared a video available at \url{https://youtu.be/u32Xnz2hDkE} showing the learning progress and neural activity over the Caltech face and motorbike task. Here we presented the face and motorbike training examples,
propagated the corresponding spike waves, and applied the STDP
rule. The input image is presented at the top-left corner of the screen. The output spikes of the input layer (i.e., DoG layer) at each time step is presented in the top-middle panel, and the accumulation of theses spikes is shown in the top-right panel. For each of the subsequent convolutional layers, the preferred features, the output spikes at each time step, and the accumulation of the output spikes are presented in the corresponding panels. Note that 4, 8, and 2 features from the first, second and third convolutional layers are selected and shown, respectively. As mentioned, the learning occurs layer by layer, thus, the label of the layer which is currently doing the learning is specified by the red color. As seen, the first layer learns to detect edges, the second layer learns intermediate features, and finally the third layer learns face and motorbike prototype features.

\section*{Acknowledgments}
This research received funding from the European Research Council under the European Unions 7th Framework Program (FP/2007-2013) / ERC Grant Agreement n.323711 (M4 project). The authors thank the NVIDIA Academic Programs team for donating a GPU hardware.


\begin{thebibliography}{10}

\bibitem{bengio2015towards}
Yoshua Bengio, Dong-Hyun Lee, Jorg Bornschein, and Zhouhan Lin.
\newblock Towards biologically plausible deep learning.
\newblock {\em arXiv:1502.04156}, 2015.

\bibitem{beyeler2013categorization}
Michael Beyeler, Nikil~D Dutt, and Jeffrey~L Krichmar.
\newblock Categorization and decision-making in a neurobiologically plausible
  spiking network using a stdp-like learning rule.
\newblock {\em Neural Networks}, 48:109--124, 2013.

\bibitem{brader2007learning}
Joseph~M Brader, Walter Senn, and Stefano Fusi.
\newblock Learning real-world stimuli in a neural network with spike-driven
  synaptic dynamics.
\newblock {\em Neural computation}, 19(11):2881--2912, 2007.

\bibitem{burbank2015mirrored}
Kendra~S Burbank.
\newblock Mirrored stdp implements autoencoder learning in a network of spiking
  neurons.
\newblock {\em PLoS Computational Biology}, 11(12):e1004566, 2015.

\bibitem{cadieu2014deep}
Charles~F Cadieu, Ha~Hong, Daniel~LK Yamins, Nicolas Pinto, Diego Ardila,
  Ethan~A Solomon, Najib~J Majaj, and James~J DiCarlo.
\newblock Deep neural networks rival the representation of primate it cortex
  for core visual object recognition.
\newblock {\em PLoS Computational Biology}, 10(12):e1003963, 2014.

\bibitem{cao2015spiking}
Yongqiang Cao, Yang Chen, and Deepak Khosla.
\newblock Spiking deep convolutional neural networks for energy-efficient
  object recognition.
\newblock {\em International Journal of Computer Vision}, 113(1):54--66, 2015.

\bibitem{cichy2016comparison}
Radoslaw~Martin Cichy, Aditya Khosla, Dimitrios Pantazis, Antonio Torralba, and
  Aude Oliva.
\newblock Comparison of deep neural networks to spatio-temporal cortical
  dynamics of human visual object recognition reveals hierarchical
  correspondence.
\newblock {\em Scientific Reports}, 6, 2016.

\bibitem{delorme2001networks}
Arnaud Delorme, Laurent Perrinet, and Simon~J Thorpe.
\newblock Networks of integrate-and-fire neurons using rank order coding b:
  Spike timing dependent plasticity and emergence of orientation selectivity.
\newblock {\em Neurocomputing}, 38:539--545, 2001.

\bibitem{dicarlo2007untangling}
James~J DiCarlo and David~D Cox.
\newblock Untangling invariant object recognition.
\newblock {\em Trends in Cognitive Sciences}, 11(8):333--341, 2007.

\bibitem{dicarlo2012does}
James~J DiCarlo, Davide Zoccolan, and Nicole~C Rust.
\newblock How does the brain solve visual object recognition?
\newblock {\em Neuron}, 73(3):415--434, 2012.

\bibitem{diehl2015unsupervised}
Peter~U Diehl and Matthew Cook.
\newblock Unsupervised learning of digit recognition using
  spike-timing-dependent plasticity.
\newblock {\em Frontiers in computational neuroscience}, 9:99, 2015.

\bibitem{diehl2015fast}
Peter~U Diehl, Daniel Neil, Jonathan Binas, Matthew Cook, Shih-Chii Liu, and
  Michael Pfeiffer.
\newblock Fast-classifying, high-accuracy spiking deep networks through weight
  and threshold balancing.
\newblock In {\em IEEE International Joint Conference on Neural Networks
  (IJCNN)}, pages 1--8, Killarney, Ireland, July 2015.

\bibitem{diehl2016conversion}
Peter~U Diehl, Guido Zarrella, Andrew Cassidy, Bruno~U Pedroni, and Emre
  Neftci.
\newblock Conversion of artificial recurrent neural networks to spiking neural
  networks for low-power neuromorphic hardware.
\newblock In {\em IEEE International Conference on Rebooting Computing}, pages
  1--8, San Diego, California, USA, October 2016.

\bibitem{doya2000complementary}
Kenji Doya.
\newblock Complementary roles of basal ganglia and cerebellum in learning and
  motor control.
\newblock {\em Current Opinion in Neurobiology}, 10(6):732--739, 2000.

\bibitem{Fukushima1980}
K~Fukushima.
\newblock Neocognitron : a self organizing neural network model for a mechanism
  of pattern recognition unaffected by shift in position.
\newblock {\em Biological Cybernetics}, 36(4):193--202, 1980.

\bibitem{ghodrati2014feedforward}
Masoud Ghodrati, Amirhossein Farzmahdi, Karim Rajaei, Reza Ebrahimpour, and
  Seyed-Mahdi Khaligh-Razavi.
\newblock Feedforward object-vision models only tolerate small image variations
  compared to human.
\newblock {\em Frontiers in Computational Neuroscience}, 8(74):1--17, 2014.

\bibitem{habenschuss2012homeostatic}
Stefan Habenschuss, Johannes Bill, and Bernhard Nessler.
\newblock Homeostatic plasticity in bayesian spiking networks as expectation
  maximization with posterior constraints.
\newblock In {\em Advances in Neural Information Processing Systems}, pages
  773--781, Lake Tahoe, Nevada, USA, December 2012.

\bibitem{huang2014associative}
Shiyong Huang, Carlos Rozas, Mario Trevi{\~n}o, Jessica Contreras, Sunggu Yang,
  Lihua Song, Takashi Yoshioka, Hey-Kyoung Lee, and Alfredo Kirkwood.
\newblock Associative hebbian synaptic plasticity in primate visual cortex.
\newblock {\em The Journal of Neuroscience}, 34(22):7575--7579, 2014.

\bibitem{hung2005fast}
Chou~P Hung, Gabriel Kreiman, Tomaso Poggio, and James~J DiCarlo.
\newblock Fast readout of object identity from macaque inferior temporal
  cortex.
\newblock {\em Science}, 310(5749):863--866, 2005.

\bibitem{hunsberger2015spiking}
Eric Hunsberger and Chris Eliasmith.
\newblock Spiking deep networks with lif neurons.
\newblock {\em arXiv:1510.08829}, 2015.

\bibitem{hussain2014improved}
Shaista Hussain, Shih-Chii Liu, and Arindam Basu.
\newblock Improved margin multi-class classification using dendritic neurons
  with morphological learning.
\newblock In {\em IEEE International Symposium on Circuits and Systems
  (ISCAS)}, pages 2640--2643, Melbourne, VIC, Australia, 2014.

\bibitem{khaligh2014deep}
Seyed-Mahdi Khaligh-Razavi and Nikolaus Kriegeskorte.
\newblock Deep supervised, but not unsupervised, models may explain it cortical
  representation.
\newblock {\em PLoS Computational Biology}, 10(11):e1003915, 2014.

\bibitem{kheradpisheh2016bio}
Saeed~Reza Kheradpisheh, Mohammad Ganjtabesh, and Timoth{\'e}e Masquelier.
\newblock Bio-inspired unsupervised learning of visual features leads to robust
  invariant object recognition.
\newblock {\em Neurocomputing}, 205:382--392, 2016.

\bibitem{kheradpisheh2015deep}
Saeed~Reza Kheradpisheh, Masoud Ghodrati, Mohammad Ganjtabesh, and Timoth{\'e}e
  Masquelier.
\newblock Deep networks resemble human feed-forward vision in invariant object
  recognition.
\newblock {\em Scientific Reports}, 6:32672, 2016.

\bibitem{kheradpisheh2016humans}
Saeed~Reza Kheradpisheh, Masoud Ghodrati, Mohammad Ganjtabesh, and Timoth{\'e}e
  Masquelier.
\newblock Humans and deep networks largely agree on which kinds of variation
  make object recognition harder.
\newblock {\em Frontiers in Computational Neuroscience}, 10:92, 2016.

\bibitem{Kirchner2006}
H~Kirchner and S~J Thorpe.
\newblock {Ultra-rapid object detection with saccadic eye movements: Visual
  processing speed revisited}.
\newblock {\em Vision Research}, 46(11):1762--1776, 2006.

\bibitem{Krizhevsky2012}
Alex Krizhevsky, I~Sutskever, and GE~Hinton.
\newblock Imagenet classification with deep convolutional neural networks.
\newblock In {\em Neural Information Processing Systems (NIPS)}, pages 1--9,
  Lake Tahoe, Nevada, USA, 2012.

\bibitem{LeCun1998}
Y~LeCun and Y~Bengio.
\newblock Convolutional networks for images, speech, and time series.
\newblock In {\em The Handbook of Brain Theory and Neural Networks}, pages
  255--258. Cambridge, MA: MIT Press, 1998.

\bibitem{lecun2015deep}
Yann LeCun, Yoshua Bengio, and Geoffrey Hinton.
\newblock Deep learning.
\newblock {\em Nature}, 521(7553):436--444, 2015.

\bibitem{lecun1998gradient}
Yann LeCun, L{\'e}on Bottou, Yoshua Bengio, and Patrick Haffner.
\newblock Gradient-based learning applied to document recognition.
\newblock {\em Proceedings of the IEEE}, 86(11):2278--2324, 1998.

\bibitem{Lee2009}
Honglak Lee, Roger Grosse, Rajesh Ranganath, and Andrew~Y. Ng.
\newblock Convolutional deep belief networks for scalable unsupervised learning
  of hierarchical representations.
\newblock pages 1--8, New York, New York, USA, 2009. ACM Press.

\bibitem{Lichtsteiner2007}
P~Lichtsteiner, C~Posch, and T~Delbruck.
\newblock {An 128x128 120dB 15us-latency temporal contrast vision sensor}.
\newblock {\em IEEE J. Solid State Circuits}, 43(2):566--576, 2007.

\bibitem{liu2009timing}
Hesheng Liu, Yigal Agam, Joseph~R Madsen, and Gabriel Kreiman.
\newblock Timing, timing, timing: fast decoding of object information from
  intracranial field potentials in human visual cortex.
\newblock {\em Neuron}, 62(2):281--290, 2009.

\bibitem{Maass2002}
Wolfgang Maass.
\newblock Computing with spikes.
\newblock {\em Special Issue on Foundations of Information Processing of
  TELEMATIK}, 8(1):32--36, 2002.

\bibitem{Martinez-Canada2016}
Pablo Mart{\'{i}}nez-Ca{\~{n}}ada, Christian Morillas, Bego{\~{n}}a Pino,
  Eduardo Ros, and Francisco Pelayo.
\newblock {A Computational Framework for Realistic Retina Modeling}.
\newblock {\em International Journal of Neural Systems}, 26(7):1650030, 2016.

\bibitem{masquelier2007unsupervised}
Timoth{\'e}e Masquelier and Simon~J Thorpe.
\newblock Unsupervised learning of visual features through spike timing
  dependent plasticity.
\newblock {\em PLoS Computational Biology}, 3(2):e31, 2007.

\bibitem{mcmahon2012stimulus}
David~BT McMahon and David~A Leopold.
\newblock Stimulus timing-dependent plasticity in high-level vision.
\newblock {\em Current Biology}, 22(4):332--337, 2012.

\bibitem{meliza2006receptive}
C~Daniel Meliza and Yang Dan.
\newblock Receptive-field modification in rat visual cortex induced by paired
  visual stimulation and single-cell spiking.
\newblock {\em Neuron}, 49(2):183--189, 2006.

\bibitem{o2015real}
Peter O'Connor, Daniel Neil, Shih-Chii Liu, Tobi Delbruck, and Michael
  Pfeiffer.
\newblock Real-time classification and sensor fusion with a spiking deep belief
  network.
\newblock {\em Frontiers in Neuroscience}, 7:178, 2013.

\bibitem{panda2016unsupervised}
Priyadarshini Panda and Kaushik Roy.
\newblock Unsupervised regenerative learning of hierarchical features in
  spiking deep networks for object recognition.
\newblock In {\em IEEE International Joint Conference on Neural Networks
  (IJCNN)}, pages 1--8, Vancouver, Canada, July 2016.

\bibitem{pignatelli2015role}
Marco Pignatelli and Antonello Bonci.
\newblock Role of dopamine neurons in reward and aversion: a synaptic
  plasticity perspective.
\newblock {\em Neuron}, 86(5):1145--1157, 2015.

\bibitem{pinto2011comparing}
Nicolas Pinto, Youssef Barhomi, David~D Cox, and James~J DiCarlo.
\newblock Comparing state-of-the-art visual features on invariant object
  recognition tasks.
\newblock In {\em IEEE workshop on Applications of Computer Vision (WACV)},
  pages 463--470, Kona, Hawaii, USA, 2011.

\bibitem{portelli2016rank}
Geoffrey Portelli, John~M Barrett, Gerrit Hilgen, Timoth{\'e}e Masquelier,
  Alessandro Maccione, Stefano Di~Marco, Luca Berdondini, Pierre Kornprobst,
  and Evelyne Sernagor.
\newblock Rank order coding: a retinal information decoding strategy revealed
  by large-scale multielectrode array retinal recordings.
\newblock {\em Eneuro}, 3(3):ENEURO--0134, 2016.

\bibitem{querlioz2013immunity}
Damien Querlioz, Olivier Bichler, Philippe Dollfus, and Christian Gamrat.
\newblock Immunity to device variations in a spiking neural network with
  memristive nanodevices.
\newblock {\em IEEE Transactions on Nanotechnology}, 12(3):288--295, 2013.

\bibitem{edmund2002computational}
Edmund~T.. Rolls and Gustavo Deco.
\newblock {\em Computational neuroscience of vision}.
\newblock Oxford university press, Oxford, UK, 2002.

\bibitem{rousselet2003taking}
Guillaume~A Rousselet, Simon~J Thorpe, and Mich{\`e}le Fabre-Thorpe.
\newblock Taking the max from neuronal responses.
\newblock {\em Trends in Cognitive Sciences}, 7(3):99--102, 2003.

\bibitem{Serrano-Gotarredona2013}
T~Serrano-Gotarredona, T~Masquelier, T~Prodromakis, G~Indiveri, and
  B~Linares-Barranco.
\newblock {STDP and STDP variations with memristors for spiking neuromorphic
  learning systems.}
\newblock {\em Frontiers in neuroscience}, 7(February):2, jan 2013.

\bibitem{Serre2007.PAMI}
T~Serre, L~Wolf, S~Bileschi, M~Riesenhuber, and T~Poggio.
\newblock Robust object recognition with cortex-like mechanisms.
\newblock {\em IEEE Transactions on Pattern Analysis Machine Intelligence},
  29(3):411--426, 2007.

\bibitem{serre2007feedforward}
Thomas Serre, Aude Oliva, and Tomaso Poggio.
\newblock A feedforward architecture accounts for rapid categorization.
\newblock {\em Proceedings of the National Academy of Sciences},
  104(15):6424--6429, 2007.

\bibitem{shoham2006silent}
Shy Shoham, Daniel~H O’Connor, and Ronen Segev.
\newblock How silent is the brain: is there a “dark matter” problem in
  neuroscience?
\newblock {\em Journal of Comparative Physiology A}, 192(8):777--784, 2006.

\bibitem{simonyan2014very}
Karen Simonyan and Andrew Zisserman.
\newblock Very deep convolutional networks for large-scale image recognition.
\newblock {\em arXiv:1409.1556}, 2014.

\bibitem{Sivilotti1991}
M~Sivilotti.
\newblock {\em {Wiring considerations in analog VLSI systems with application
  to field-programmable networks}}.
\newblock PhD thesis, Comput. Sci. Div., California Inst. Technol., Pasadena,
  CA, 1991.

\bibitem{thorpe2001spike}
Simon Thorpe, Arnaud Delorme, and Rufin Van~Rullen.
\newblock Spike-based strategies for rapid processing.
\newblock {\em Neural Networks}, 14(6):715--725, 2001.

\bibitem{thorpe1996speed}
Simon Thorpe, Denis Fize, Catherine Marlot, et~al.
\newblock Speed of processing in the human visual system.
\newblock {\em Nature}, 381(6582):520--522, 1996.

\bibitem{van2001rate}
Rufin Van~Rullen and Simon~J Thorpe.
\newblock Rate coding versus temporal order coding: what the retinal ganglion
  cells tell the visual cortex.
\newblock {\em Neural Computation}, 13(6):1255--1283, 2001.

\bibitem{Wohrer2009}
Adrien Wohrer and Pierre Kornprobst.
\newblock {Virtual Retina: a biological retina model and simulator, with
  contrast gain control.}
\newblock 26(2):219--49, apr 2009.

\bibitem{Yousefzadeh2015}
A.~Yousefzadeh, T.~Serrano-Gotarredona, and B.~Linares-Barranco.
\newblock {Fast Pipeline 128??128 pixel spiking convolution core for
  event-driven vision processing in FPGAs}.
\newblock In {\em Proceedings of 1st International Conference on Event-Based
  Control, Communication and Signal Processing, EBCCSP 2015}, 2015.

\bibitem{zeiler2014visualizing}
Matthew~D Zeiler and Rob Fergus.
\newblock Visualizing and understanding convolutional networks.
\newblock In {\em European Conference on Computer Vision (ECCV)}, pages
  818--833, Zurich, Switzerland, September 2014.

\bibitem{zhao2015feedforward}
Bo~Zhao, Ruoxi Ding, Shoushun Chen, Bernabe Linares-Barranco, and Huajin Tang.
\newblock Feedforward categorization on aer motion events using cortex-like
  features in a spiking neural network.
\newblock {\em IEEE Transactions on Neural Networks and Learning Systems},
  26(9):1963--1978, 2015.

\end{thebibliography}







\end{document}